\documentclass[10pt,twocolumn,letterpaper]{article}
\usepackage{soul}
\usepackage{booktabs} 
\usepackage{graphicx}
\usepackage[table]{xcolor}
 \usepackage{cvpr}              

\definecolor{cvprblue}{rgb}{0.21,0.49,0.74}
\usepackage[pagebackref,breaklinks,colorlinks,allcolors=cvprblue]{hyperref}

\colorlet{myblue}{cyan!40}
\definecolor{codegreen}{rgb}{0,0.6,0}

\definecolor{codegreen}{rgb}{0,0.6,0}
\usepackage{wrapfig}

\usepackage[floatrowsep=quad]{floatrow}

\usepackage{float}
\usepackage{algorithm}
\usepackage{algpseudocode}

\usepackage{dsfont}
\usepackage{multirow}
\usepackage{pifont}
\usepackage{stfloats}
\usepackage{subfig}
\usepackage{caption}

\usepackage{mdframed}

\def\paperID{0000} 
\def\confName{CVPR}
\def\confYear{2025}

\title{
    \begin{minipage}{0.125\textwidth}
        \centering
        \includegraphics[width=\textwidth]{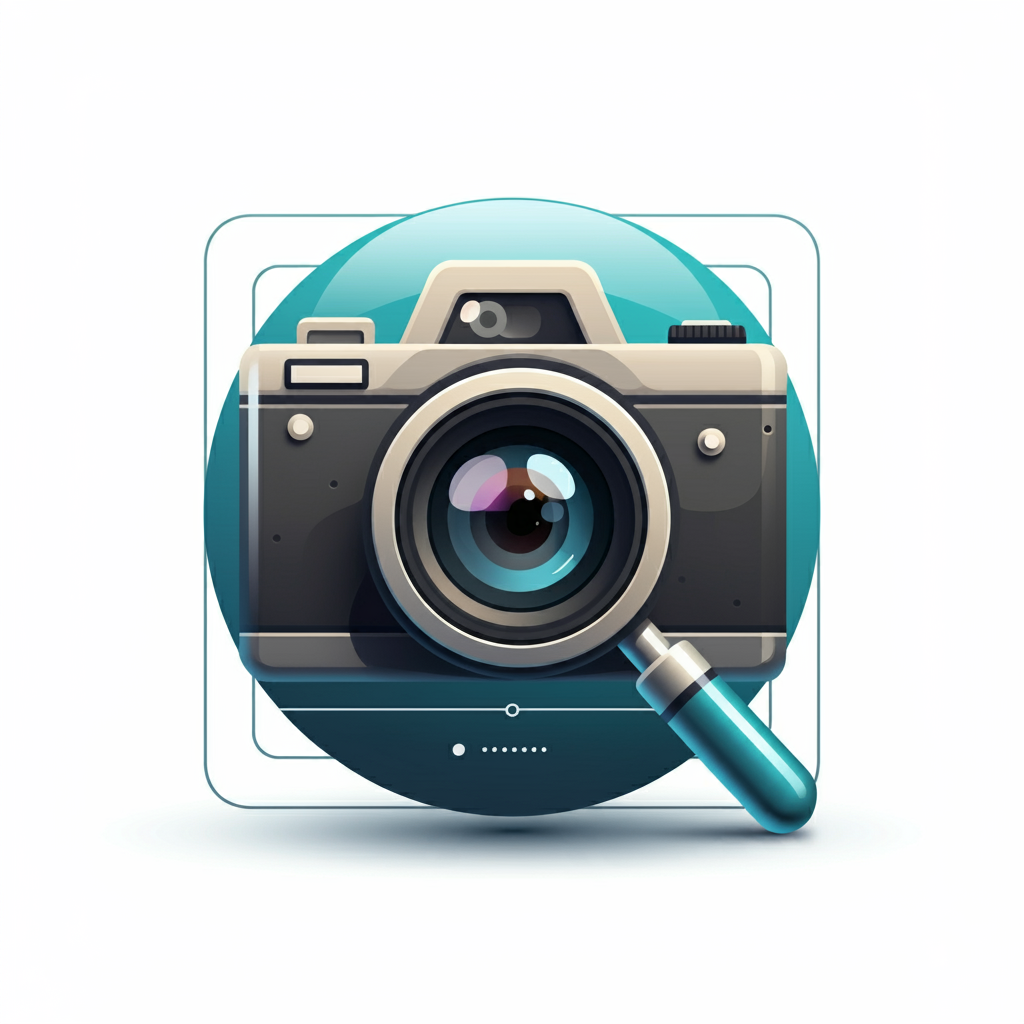}
    \end{minipage}
    \hspace{0.5em} 
    \begin{minipage}{0.85\textwidth}
        \centering
        \underline{TruthLens} \\ 
        A Training-Free Paradigm for DeepFake Detection
    \end{minipage}
}

\author{Ritabrata Chakraborty\\
Manipal University Jaipur\\
{\tt\small ritabrata.229301716@muj.manipal.edu}
\and
Rajatsubhra Chakraborty\\
University of North Carolina Charlotte\\
{\tt\small rchakra6@charlotte.edu}
\and
Ali Khaleghi Rahimian\\
University of North Carolina Charlotte\\
{\tt\small akhalegh@charlotte.edu}
\and
Thomas MacDougall\\
University of North Carolina Charlotte\\
{\tt\small tmacdoug@charlotte.edu}
}

\begin{document}

\maketitle
\begin{abstract}
The proliferation of synthetic images generated by advanced AI models poses significant challenges in identifying and understanding manipulated visual content. Current fake image detection methods predominantly rely on binary classification models that focus on accuracy while often neglecting interpretability, leaving users without clear insights into why an image is deemed real or fake. To bridge this gap, we introduce \textbf{TruthLens}, a novel training-free framework that reimagines deepfake detection as a visual question-answering (VQA) task. TruthLens utilizes state-of-the-art large vision-language models (LVLMs) to observe and describe visual artifacts and combines this with the reasoning capabilities of large language models (LLMs) like GPT-4 to analyze and aggregate evidence into informed decisions. By adopting a multimodal approach, TruthLens seamlessly integrates visual and semantic reasoning to not only classify images as real or fake but also provide interpretable explanations for its decisions. This transparency enhances trust and provides valuable insights into the artifacts that signal synthetic content. Extensive evaluations demonstrate that TruthLens outperforms conventional methods, achieving high accuracy on challenging datasets while maintaining a strong emphasis on explainability. By reframing deepfake detection as a reasoning-driven process, TruthLens establishes a new paradigm in combating synthetic media, combining cutting-edge performance with interpretability to address the growing threats of visual disinformation.

\end{abstract}

\section{Introduction}\label{sec:introduction}

The proliferation of manipulated and synthetic images, driven by advancements in generative models such as Generative Adversarial Networks (GANs) and diffusion models, has created significant challenges in distinguishing real from fake images. Tools like StyleGAN \cite{karras2019stylegan} and recent diffusion models \cite{ho2020denoising} have enabled the creation of highly photorealistic images, which are increasingly used in contexts ranging from entertainment to malicious disinformation campaigns. This surge has raised critical concerns in domains such as media integrity, cybersecurity, and ethical AI.


Traditional methods for detecting fake images rely heavily on binary classifiers, such as CNNDetection \cite{cnn_detection}, which uses pre-trained Convolutional Neural Networks (CNNs) like ResNet-50 to identify specific artifacts in GAN-generated images. While effective for early GAN models, these methods struggle with newer architectures that exhibit fewer detectable flaws. More recently, approaches like Diffusion Reconstruction Error (DIRE) \cite{dire} have shown promise by leveraging the reconstruction inconsistencies of diffusion models to detect synthetic content. However, these methods focus solely on classification, offering little interpretability or insight into why an image is labeled as fake.

This lack of interpretability is problematic, particularly in high-stakes applications where trust and accountability are paramount. Recent studies suggest that incorporating reasoning into classification tasks can enhance transparency and user trust~\cite{ribeiro2016should, selvaraju2017grad}. For instance, Ribeiro et al.~\cite{ribeiro2016should} proposed the LIME framework, which provides local explanations for machine learning predictions. Similarly, methods like Grad-CAM \cite{selvaraju2017grad} visualize important regions in images for CNN-based classifiers, but they are not inherently designed to address the complexities of synthetic image detection.

Inspired by advances in Large Vision-Language Models (LVLMs), such as LLaVA \cite{llava} and BLIP-2 \cite{blip2}, we propose re-framing the task of fake image detection as a multimodal Visual Question Answering (VQA) problem. In this paper, we introduce TruthLens, a LVLM+LLM framework that bridges the gap between detection accuracy and interpretability in fake image detection. By combining the detection capabilities of traditional models with the reasoning power of multimodal systems, TruthLens not only classifies images as real or fake but also provides detailed justifications for its decisions by leveraging both visual and textual features. The framework incorporates a structured pipeline that integrates multimodal querying, textual aggregation, and reasoning to deliver transparent and robust results.

The contributions of TruthLens are twofold:

\begin{enumerate}
    \item \textbf{Multimodal Reasoning for Enhanced Interpretability:} TruthLens reframes fake image detection as a Visual Question Answering (VQA) problem, utilizing Large Vision-Language Models (LVLMs) to integrate visual and textual modalities. This approach enables the identification of key features or artifacts in images, providing meaningful, natural language justifications for classification decisions.

    \item \textbf{Structured Pipeline for Transparent Decision-Making:} The framework employs a structured pipeline that includes multimodal querying, textual aggregation, and final reasoning. It organizes outputs into concise, human-readable summaries and utilizes a language model to classify images as real or fake, offering detailed, interpretable explanations for its conclusions.
\end{enumerate}

This paper is organized as follows: Section \ref{sec:related_work} provides a review of related work in fake image detection and multimodal reasoning. Section \ref{sec:methodology} describes the proposed framework in detail. Section \ref{sec:experiments} presents the experimental results, and Section \ref{sec:conclusion} concludes with key insights and potential future directions.

\section{Related Work}\label{sec:related_work}

The task of detecting synthetic images has evolved significantly with advancements in generative models. Early methods focused on binary classification using handcrafted features, while modern approaches leverage deep learning models, which excel at uncovering subtle artifacts in fake images. Additionally, advancements in multimodal systems, large language models, and vision-language models have opened new avenues for interpretable fake image detection.

\subsection{Large Language Models}
Large Language Models (LLMs) are characterized by their massive number of parameters and large-scale training corpora. Models such as GPT \cite{brown2020languagemodelsfewshotlearners} and LLaMA \cite{touvron2023llamaopenefficientfoundation} exemplify this class of models and are renowned for their versatility and power. Using transformer architectures, LLMs process language in a manner that surpasses traditional methods, achieving human-like or near-human performance in natural language tasks.

Many LLMs, like GPT \cite{brown2020languagemodelsfewshotlearners}, are so large, that they can almost be considered a distillation of nearly all human knowledge. This size needs the support of a large training corpus, which is typically sourced from all across the internet. With such large models and training corpus, LLMs are considered to operate at a near-human level in ideal conditions. Additionally, the number of tasks they can be generalized to is quite high. Nowadays, many vision models make use of pre-trained LLMs in order to expand their own capabilities, as there is a great amount of overlap between many vision and language tasks.

\subsection{Large Vision-Language Models}
LLMs are not limited to language tasks; they have also been adapted to handle computer vision tasks. Vision-Language Models (VLMs) combine visual and textual encoders within a unified architecture, allowing them to perform both categories of tasks simultaneously. Notable examples include OpenAI's GPT-for-vision (GPT4V), which processes both images and text for multi-modal reasoning, and LLaVA \cite{llava}, which integrates several powerful vision and language models to enhance understanding of complex instruction-following tasks. These models are well-known for their generalization capabilities and broad applicability, proving particularly effective for reasoning over multi-modal inputs.

Some notable models include Chat-UniVi \cite{jin2024chatuniviunifiedvisualrepresentation}, which merges image and video tokens into a single unified representation. This merging provides Chat-UniVi with the ability to understand both forms of media, and enhances its LLM capabilities \cite{jin2024chatuniviunifiedvisualrepresentation}. Other models, like BLIP-2 \cite{blip2} take a different approach to improving model capabilities, and focus on different pre-training strategies in order to teach the model better multi-modal representations of data. BLIP-2 specifically learns in two stages: a representation learning stage, using a frozen image encoder, and a generative learning stage, using a frozen LLM \cite{blip2}. This "bootstrapping" of pre-trained models allows BLIP-2 to take advantage of powerful vision and language models, and combine their capabilities into one. Finally, CogVLM \cite{wang2024cogvlmvisualexpertpretrained} modifies the typical frozen language and vision encoders, by adding an extra "expert layer" \cite{wang2024cogvlmvisualexpertpretrained} inside the transformers. This extra layer is meant to connect the two normally separate encoders, integrating both visual and linguistic features into one.

\subsection{Deepfakes and Deepfake Detection}
The artificial generation and manipulation of human faces—commonly referred to as "deepfakes"—gained prominence with the release of StyleGAN \cite{karras2019stylegan}. These deepfakes have grown increasingly realistic over time, making their detection a critical research area. 

Early detection methods, such as CNNDetection \cite{cnn_detection}, employed pre-trained CNNs like ResNet-50 to classify real and fake images based on pixel-level inconsistencies. While effective for earlier GANs, the flaws exploited by these methods are less prevalent in modern GANs and diffusion-based image generation.

Diffusion Reconstruction Error (DIRE) \cite{dire} was introduced as an alternative to binary classifiers. This method reconstructs input images using pre-trained diffusion models and calculates the difference between the input and reconstructed image. DIRE assumes that diffusion-generated images share a similar probability distribution, allowing it to effectively detect diffusion-based fakes. However, it struggles with images generated by models outside this distribution.

\subsection{Datasets for Deepfake Detection}

Several benchmark datasets have been developed to support the training and evaluation of deepfake detection models. \textit{CIFAKE} \cite{cifake} combines real and synthetic images derived from the CIFAR-10 dataset, providing a compact yet challenging benchmark. \textit{CelebA-HQ Resized} \cite{karras2018progressive} features high-resolution images of celebrity faces, making it particularly useful for evaluating generative models. Furthermore, benchmarks like \textit{FakeBench} \cite{fakebench} include datasets such as FakeClass and FakeClue, designed to assess both detection accuracy and reasoning capabilities. Another notable dataset, \textit{FakeQA}, offers over 40,000 question-answer pairs, enabling open-ended evaluation of multi-modal models in reasoning and detection tasks. Lastly, \textit{ForgeryNet} \cite{forgerynet} provides a large-scale dataset with over 2.9 million images and videos, encompassing diverse manipulation techniques to facilitate robust and comprehensive evaluations.

\subsection{Interpretability in Machine Learning}
Interpretability has become a crucial aspect in fostering trust in AI systems, particularly in the domain of deepfake detection. While general-purpose frameworks like LIME \cite{ribeiro2016should} and Grad-CAM \cite{selvaraju2017grad} have proven effective in explaining model predictions across various tasks, there is a growing need for specialized interpretability methods tailored to the nuances of deepfake analysis.

Recent research has explored the potential of Large Language Models (LLMs) in detecting deepfakes. For instance, studies have investigated whether LLMs can distinguish between real and AI-generated content, leveraging their broad knowledge and contextual understanding \cite{zellers2023llms}. These approaches aim to complement traditional image-based detection methods by analyzing textual and semantic inconsistencies that may be present in deepfake content.

In parallel, initiatives like FAKEBENCH have emerged to provide comprehensive evaluation frameworks for deepfake detection algorithms \cite{wang2023fakebench}. FAKEBENCH offers a standardized platform for assessing the performance of various detection methods, including those that incorporate interpretability components. This benchmark not only evaluates detection accuracy but also considers the explainability of the models, addressing the critical need for transparent and trustworthy AI systems in combating digital misinformation.

The development of interpretable deepfake detection models presents unique challenges due to the sophisticated nature of modern forgery techniques \cite{li2023interpretable}. Researchers are working on adapting existing explainability methods and developing new ones that can effectively highlight the subtle artifacts and inconsistencies that characterize deepfakes \cite{chen2022towards}. These specialized approaches aim to provide more precise and relevant explanations for deepfake detection decisions, potentially improving both the accuracy and trustworthiness of detection systems.

\subsection{Large Vision-Language Models for Deepfake Detection}
Recent advancements in vision-language models have demonstrated their potential for combining detection and reasoning. Models like BLIP-2 \cite{blip2} and LLaVA \cite{llava} excel at joint visual and textual understanding. FakeBench \cite{fakebench} explores their applicability to deepfake detection, evaluating not only detection accuracy but also reasoning capabilities. However, these works primarily focus on benchmarking rather than developing end-to-end systems for detection and reasoning.

\subsection{Our Contribution}
Building on these advancements, our proposed framework, TruthLens, integrates detection and interpretability by re-framing deepfake detection as a Visual Question Answering (VQA) task. By leveraging state-of-the-art vision-language models, TruthLens provides both accurate classification and detailed reasoning for image authenticity, addressing limitations in existing methods.

\section{Methodology}\label{sec:methodology}

The task of detecting fake images has traditionally been approached as a binary classification problem, where the focus is solely on distinguishing between real and fake images. While these methods can achieve high accuracy, they fail to provide insights into the reasoning behind their predictions, which is critical for building trust and transparency in AI systems. With the advent of Large Vision-Language Models (LVLMs) and Large Language Models (LLMs), we propose a novel approach that re-frames fake image detection as a \textit{Visual Question Answering (VQA)} problem. This paradigm shift enables both detection and interpretability by incorporating structured reasoning into the decision-making process.

Our framework, TruthLens, leverages multimodal reasoning to classify images and provide detailed justifications for its decisions. The pipeline consists of four main steps: (1) Question Generation, (2) Multimodal Reasoning, (3) Textual Aggregation, and (4) Final Decision Making.

\subsection{Step 1: Question Generation}

The first step in the pipeline involves generating a set of predefined prompts or questions that address specific visual and textual cues in an image. These prompts are carefully designed to probe various aspects of image authenticity, such as artifacts, inconsistencies, or visual features commonly associated with synthetic images. The set of prompts, denoted as \(P = \{p_1, p_2, \dots, p_N\}\), consists of \(N\) individual prompts, each corresponding to a specific artifact or visual clue. Prompts are crafted based on known patterns in synthetic images, such as lighting inconsistencies, unnatural textures, or boundary artifacts.

By systematically querying the input image \(I\) with these prompts, the framework aims to extract detailed responses that highlight evidence supporting the classification task.

The prompts used are listed as follows:
\begin{itemize}
\item \textbf{Lighting and Shadows:} \textit{"Describe the lighting in the image. Does it appear natural or does it show any inconsistencies, such as unrealistic shadows or lighting direction?"}
\item \textbf{Texture and Skin Details:} \textit{"Analyze the texture of the skin in this image. Does the skin appear to have natural imperfections like pores, wrinkles, or blemishes, or is it unnaturally smooth?"}
\item \textbf{Symmetry and Proportions:} \textit{"Describe the facial symmetry in the image. Are there any noticeable asymmetries in the eyes, nose, mouth, or face shape?"}
\item \textbf{Reflections and Highlights:} \textit{"Examine the reflections in the eyes or any shiny areas on the skin. Do they appear to be consistent with the environment, or do they seem artificial or inconsistent?"}
\item \textbf{Facial Features and Expression:} \textit{"Describe the facial expression in the image. Does it appear natural, or are there any signs of a forced or unnatural expression?"}
\item \textbf{Facial Hair (if applicable):} \textit{"If there is facial hair in the image, describe its appearance. Does it seem realistic in terms of texture, growth pattern, and interaction with the lighting?"}
\item \textbf{Eyes and Pupils:} \textit{"Describe the appearance of the eyes in the image. Do the pupils appear natural in size, shape, and positioning, or are there any abnormalities?"}
\item \textbf{Background and Depth Perception:} \textit{"Describe the background of the image. Does it seem well-integrated with the face in terms of depth, focus, and lighting, or does it appear artificially blurred or detached?"}
\item \textbf{Overall Realism of the Face:} \textit{"Taking into account the lighting, texture, symmetry, and other features, describe the overall realism of the face. Does it show any signs of being digitally manipulated or generated?"}
\end{itemize}

These prompts are designed to capture flaws and artifacts typical to most deepfakes as best as possible. Unlike GAN-generated images, most fake images nowadays do not share the underlying artifacts that make them easy to spot \cite{cnn_detection}, and so we must prompt the model to consider more global visual features. Modern day deepfakes are much more sophisticated than they once were, but still have several common visual abnormalities that give them away: errors in texture, lighting, and anatomy are still commonplace in many current day deepfakes \cite{kamali2024distinguishaigeneratedimagesauthentic}. Prompting the model to focus on these visual abnormalities will give it the best chance of detecting fake images.

\subsection{Step 2: Multimodal Reasoning}

In the second step, a multimodal model, denoted as \(f_{\text{MM}}\), processes the input image \(I\) alongside each prompt \(p_i \in P\) to generate answers. The model combines visual and textual modalities to produce meaningful explanations. Formally, the multimodal model can be represented as:
\[
f_{\text{MM}}(I, p): \mathcal{I} \times \mathcal{P} \to \mathcal{A}
\]
where \(\mathcal{I}\) is the space of input images, \(\mathcal{P}\) is the space of textual prompts, and \(\mathcal{A}\) is the space of textual answers. For each prompt \(p_i\), the model generates an answer \(a_i = f_{\text{MM}}(I, p_i)\), resulting in a complete set of answers:
\[
A = \{a_1, a_2, \dots, a_N\}.
\]

The multimodal model extracts visual features such as textures, edges, and artifacts using vision encoders and contextualizes these features with respect to the prompt using text encoders. It then generates natural language answers combining visual and textual evidence. By leveraging state-of-the-art models like LLaVA and BLIP-2, this step ensures that the extracted answers are both precise and interpretable.

\subsection{Step 3: Textual Aggregation}

Once the answers are generated, the next step involves aggregating these responses into a structured summary \(S\) that encapsulates the key observations from all prompts. This structured summary organizes the raw outputs into a coherent explanation. The aggregation process is represented as:
\[
g: \mathcal{A}^N \to \mathcal{S}
\]
where \(\mathcal{S}\) is the space of structured summaries. The summary \(S\) is computed as:
\[
S = g(A) = g(f_{\text{MM}}(I, p_1), f_{\text{MM}}(I, p_2), \dots, f_{\text{MM}}(I, p_N)).
\]

The goal of this step is to consolidate the multimodal reasoning into a concise, human-readable format that provides the foundation for the final classification and reasoning.

\subsection{Step 4: Final Decision Making}

In the final step, the structured summary \(S\) is passed to a language model \(f_{\text{LM}}\) for classification and reasoning. The language model determines whether the image is real or fake and generates a natural language explanation for its decision. This process can be formulated as:
\[
f_{\text{LM}}(S): \mathcal{S} \to \mathcal{Y} \times \mathcal{R}
\]
where \(\mathcal{Y} = \{\text{Real, Fake}\}\) represents the space of classification labels, and \(\mathcal{R}\) represents the space of textual justifications. The final output is:
\[
(y, r) = f_{\text{LM}}(S),
\]
where \(y\) is the classification result Real or Fake and \(r\) is the explanation for the decision.

\begin{figure}[H]
    \centering
    \includegraphics[width=1\textwidth]{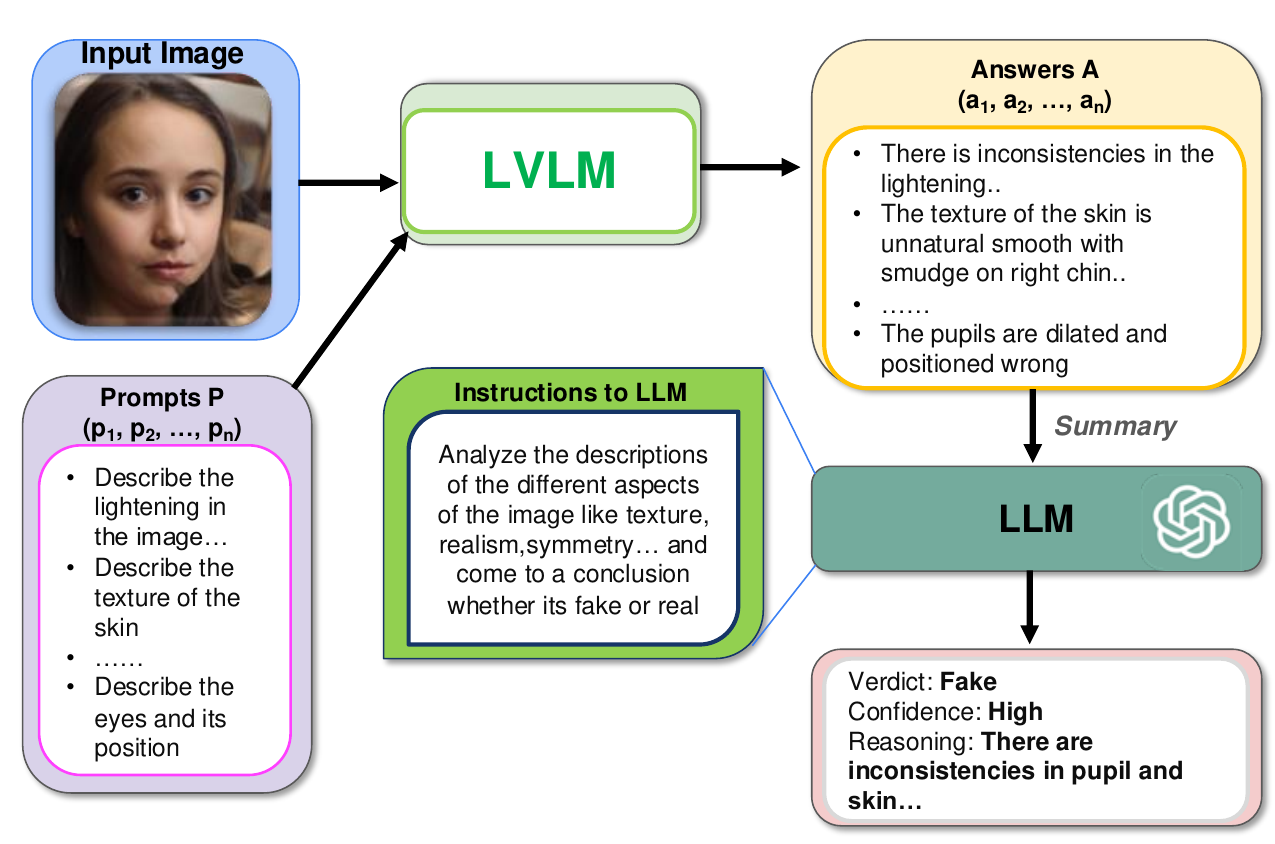}
    \caption{Overview of the detection pipeline used in the TruthLens framework.}
    \label{fig:pipeline}
\end{figure}

\section{Experiments}\label{sec:experiments}

In this section, we evaluate the performance of our proposed framework, TruthLens, on various datasets and compare it against existing state-of-the-art methods, including CNNDetection and DIRE. Our experiments are designed to measure detection accuracy, reasoning quality, and robustness across different types of fake images. 

\subsection{Datasets}
We evaluate on the following datasets:
\begin{itemize}
    \item \textbf{LDM Dataset}: Consists of 1000 fake images generated by Latent Diffusion Models (LDM) alongside corresponding 1000 real images from FFHQ~\cite{ffhq}. This dataset challenges models with high-quality synthetic images that closely mimic real-world distributions.
    \item \textbf{ProGAN Dataset}: Includes 1000 fake images generated by ProGAN derived from ForgeryNet dataset~\cite{forgerynet} . ProGAN’s images exhibit traditional GAN artifacts, making this dataset suitable for evaluating the performance of models on GAN-based generation techniques.
\end{itemize}
\begin{figure}[H]
    \centering
    \includegraphics[width=1\textwidth]{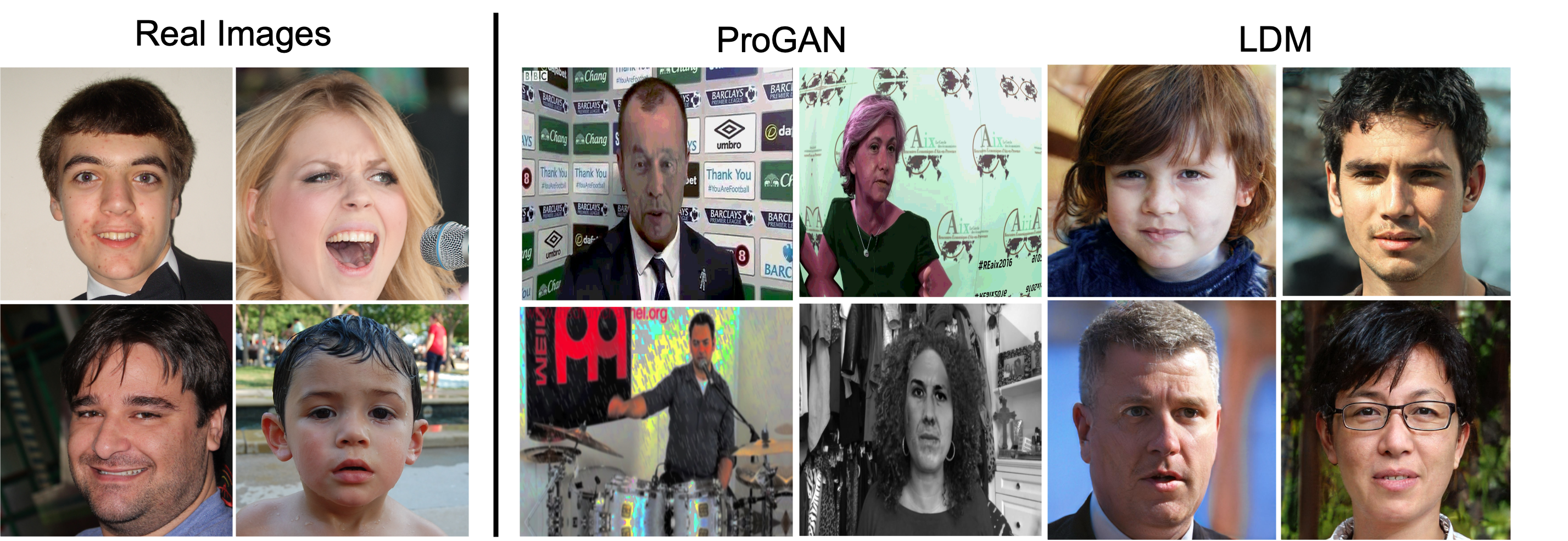}
    \caption{Overview of the evaluation dataset on the left hand side we have Real images from FFHQ dataset~\cite{ffhq} and on the right we have ProGAN generated images from ForgeryNet dataset~\cite{forgerynet} and Latent Diffusion Model(LDM)~\cite{LDM} generated images. }
    \label{fig:dataset}
\end{figure}
\subsection{Metrics}

We report results using several key metrics. Accuracy measures the percentage of correct classifications, whether the data is real or fake. AUC (Area Under the ROC Curve) indicates the model's ability to distinguish between classes. Additionally, we evaluate Precision, Recall, and F1-Score, which offer a comprehensive understanding of the balance between true positives and false negatives. Lastly, Qualitative Analysis includes visualizations and reasoning quality to assess the interpretability of the results.

\subsection{Results}

\subsubsection{Comparison of AUC Scores}
The AUC scores of various methods across LDM and ProGAN datasets are shown in Table \ref{tab:methods_comparison}. Our framework achieves superior performance compared to CNNDetection and DIRE, demonstrating its robustness across diverse generation techniques.

\begin{table}[H]
    \caption{Comparison of AUC scores across datasets generated by LDM and ProGAN.}
    \label{tab:methods_comparison}
    \centering
    \setlength{\tabcolsep}{10.7pt}
    \begin{tabular}{lcc}
        \toprule
        \textbf{Method}          & \textbf{LDM (\%)} & \textbf{ProGAN (\%)} \\
        \midrule
        DIRE                     & 46.47            & 58.12               \\
        CNNDetection             & 86.50            & 40.44               \\ \rowcolor{lightgray!40}
        TruthLens (Ours)         & \textbf{95}   & \textbf{97.5}      \\
        \bottomrule
    \end{tabular}
\end{table}

\subsubsection{Classification Accuracy with and without Prompt + LLM}
Table \ref{tab:real-fake-classification} compares the classification accuracy of different models on real and fake images (LDM and ProGAN), with and without the use of prompts and the language model (LLM). The inclusion of prompts and LLM significantly enhances detection performance, especially for challenging LDM datasets.

\begin{table}[H]
  \caption{Classification results for real and fake data (LDM and ProGAN) using different models, with and without Prompt + LLM.}
  \label{tab:real-fake-classification}
  \setlength{\tabcolsep}{2pt} 
  \renewcommand{\arraystretch}{0.8} 
  \small 
  \centering
  \begin{tabular}{ccccc}
    \toprule
    \multirow{2}{*}{Method}    & \multirow{2}{*}{Model} & \multirow{2}{*}{Real (\%)} & \multicolumn{2}{c}{Fake (\%)} \\
    \cmidrule(lr{0.5em}){4-5}
                               &                       &                       & LDM               & ProGAN             \\
    \midrule
    \multirow{4}{*}{Yes or No Question} 
                               & BLIP2            & 52                  & 20              &  50            \\
                               & CogVLM                 & 62                  & 25              & 52               \\
                               & LLAVA 1.5                 & 50                  & 22              & 48               \\
                               & ChatUniVi             & 54                  & 24              & 52               \\
    \midrule
    \multirow{4}{*}{Prompts + LLM}    
                               & BLIP2             & 74         & 68              & 72               \\
                               & CogVLM                  & 85                 & 82     &  90     \\
                               & LLAVA 1.5                  & 76                 & 70             & 74               \\
                               & ChatUniVi              & \textbf{98}                  & \textbf{92}              & \textbf{97}               \\
    \bottomrule
  \end{tabular}
\end{table}

\subsubsection{Performance Breakdown by Metric}
Table \ref{tab:metric-breakdown} provides a detailed breakdown of model performance, including precision, recall, and F1-score for LDM and ProGAN datasets. These metrics highlight the strengths of our framework in accurately identifying fake images across different generation techniques.

\begin{table}[H]
  \caption{Performance breakdown of different methods on LDM and ProGAN datasets. TruthLens demonstrates superior precision, recall, and F1-score across both datasets.}
  \label{tab:metric-breakdown}
  \setlength{\tabcolsep}{1.5pt} 
  \small 
  \centering
  \begin{tabular}{lcccccc}
    \toprule
    \multirow{2}{*}{\textbf{Method}} & \multicolumn{2}{c}{\textbf{Precision (\%)}} & \multicolumn{2}{c}{\textbf{Recall (\%)}} & \multicolumn{2}{c}{\textbf{F1-Score (\%)}} \\ 
                                     & \textbf{LDM} & \textbf{ProGAN} & \textbf{LDM} & \textbf{ProGAN} & \textbf{LDM} & \textbf{ProGAN} \\
    \midrule
    DIRE                             & 49.97       & 49.67       & \textbf{98.90}& \textbf{97.70}& \underline{66.40}& \underline{65.86}\\
    CNNDetection                     & \textbf{97.59}       & 60.00       & 8.10        & 0.30        & 14.96       & 0.60        \\ \rowcolor{lightgray!40}
    TruthLens (Ours)                 & \underline{90.99}& \textbf{90.16}& \underline{90.57}& \underline{96.28}& \textbf{95.55}& \textbf{95.91}\\
    
    \bottomrule
  \end{tabular}
\end{table}

\begin{figure}[H]
    \centering
    \includegraphics[width=0.9\textwidth, height=0.5\textwidth]{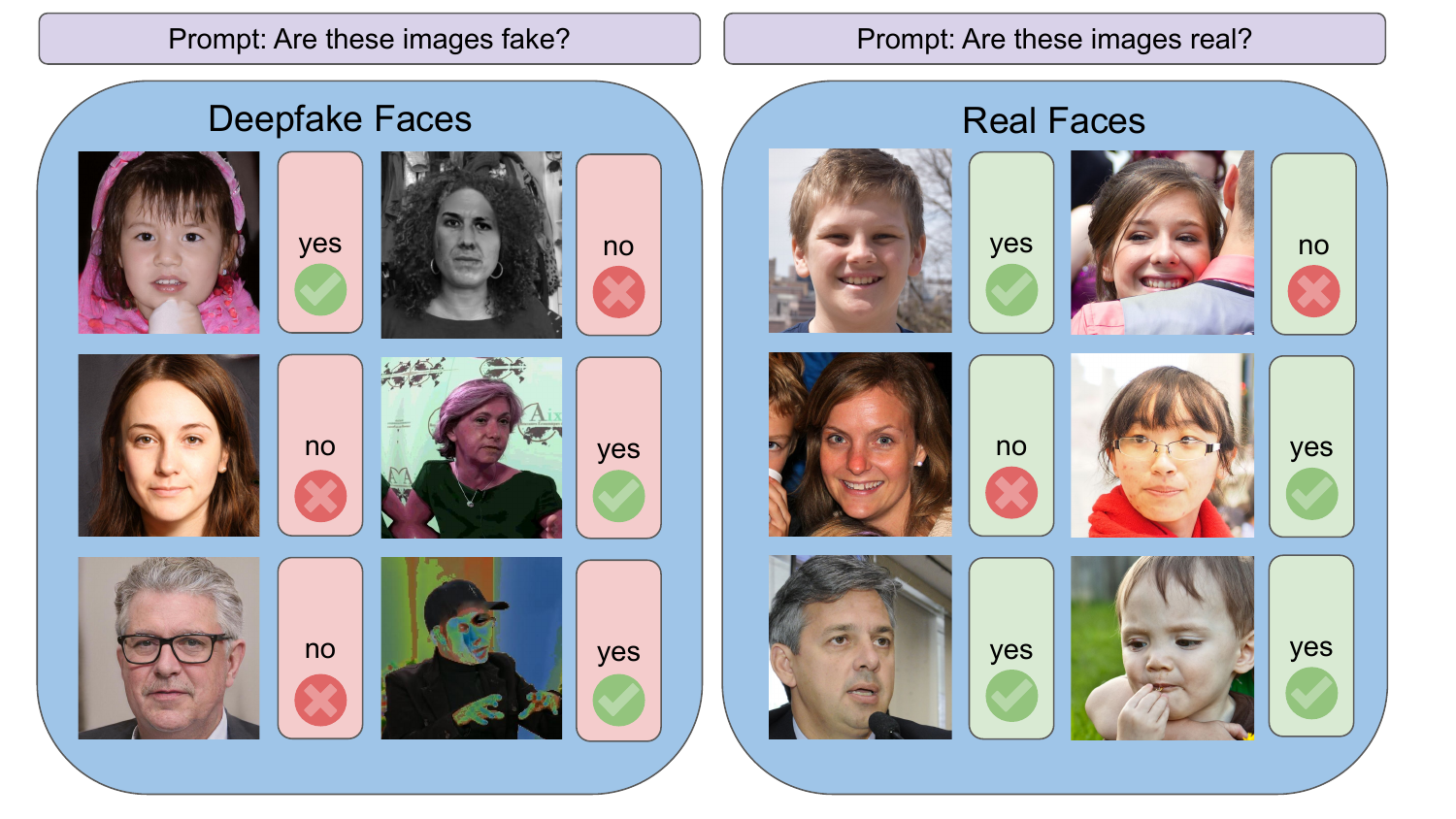} 
    \caption{A visualization of the yes/no prompts given to LVLMs, and their responses.}
    \label{fig:fake-or-real}
\end{figure}
\subsubsection{Ablation Studies}

In this section, we analyze the impact of specific image features on the detection accuracy of synthetic images through targeted ablation studies. Table \ref{tab:feature_accuracy} presents the accuracy achieved when prompts are designed to focus on distinct categories of visual cues. Each category probes a specific aspect of the image, helping to identify patterns or inconsistencies that contribute to the model's overall performance.

The results indicate that certain features, such as "Eyes and Pupils" (82.5\% accuracy) and "Facial Hair" (74.5\% accuracy), provide the most reliable cues for distinguishing real and fake images. These features are likely less prone to generative model artifacts, making them critical for accurate classification. On the other hand, categories like "Texture and Skin Details" (54.5\% accuracy) and "Overall Realism of the Face" (55.6\% accuracy) exhibit lower accuracy, suggesting these aspects are either less distinctive or more challenging for models to evaluate effectively.

The study highlights the importance of leveraging feature-specific prompts to enhance the detection process. By identifying high-impact categories, future improvements can prioritize these areas, leading to more focused and efficient detection strategies. This analysis also underscores the need for diverse and comprehensive prompts to cover a wide range of potential artifacts in synthetic images.

\begin{table}[h]
    \caption{Accuracy of Image Features across Categories.}
    \label{tab:feature_accuracy}
    \centering
    \setlength{\tabcolsep}{10.7pt}
    \begin{tabular}{lc}
        \toprule
        \textbf{Prompt Category}                    & \textbf{Accuracy (\%)} \\
        \midrule
        Lighting and Shadows                & 62.0                  \\
        Texture and Skin Details            & 54.5                  \\
        Symmetry and Proportions            & 67.8                  \\
        Reflections and Highlights          & 66.6                  \\
        Facial Features and Expression      & 67.0                  \\
        Facial Hair                         & 74.5                  \\
        Eyes and Pupils                     & 82.5                  \\
        Background and Depth Perception     & 60.0                  \\
        Overall Realism of the Face         & 55.6                  \\
        \bottomrule
    \end{tabular}
\end{table}

\subsubsection{Qualitative Analysis}
\begin{figure*}[t]
\centering
    \includegraphics[width=1\textwidth]{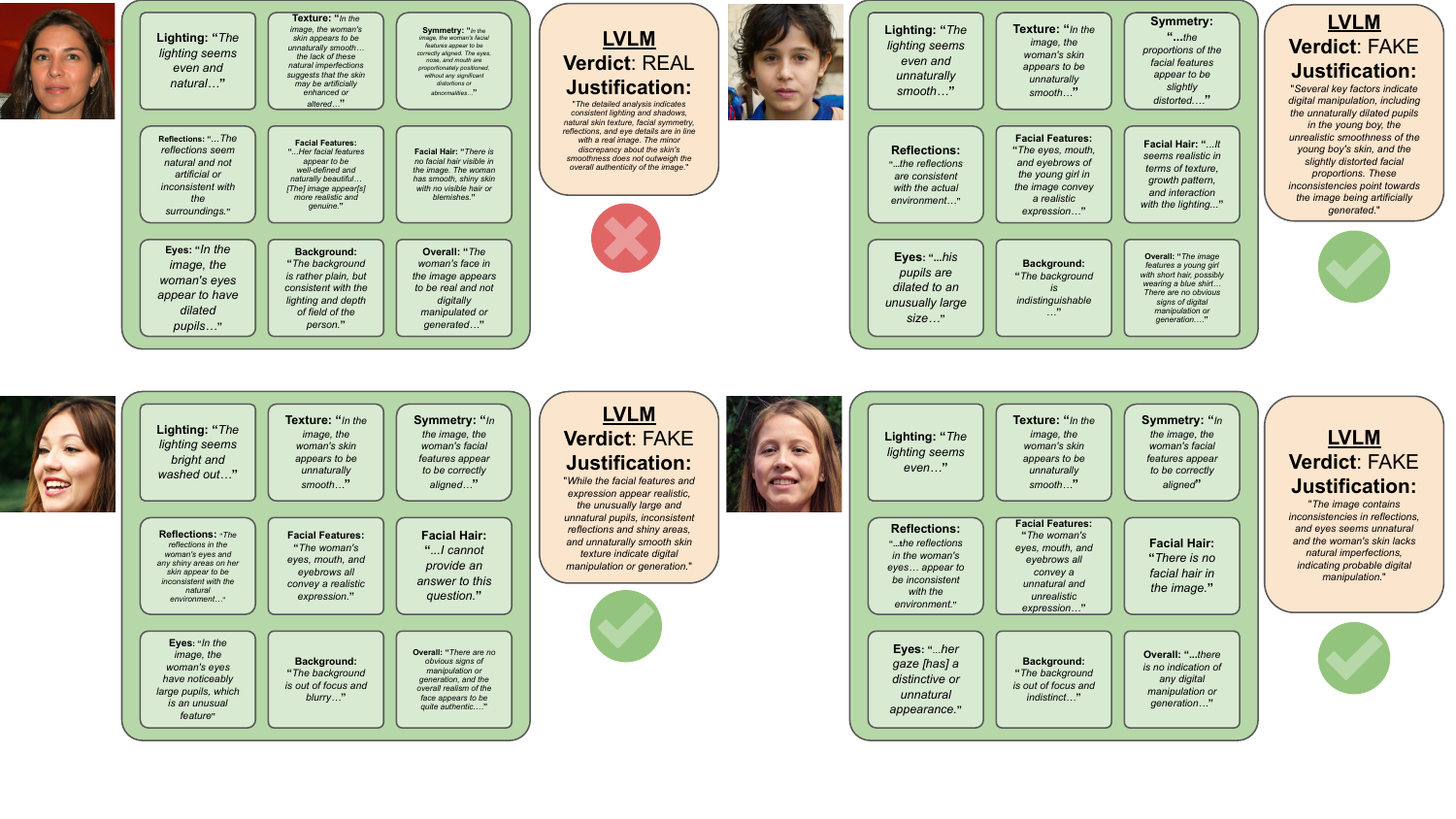}
    \caption{A visualization of the output of the model for each of the prompts, and the LVLM's final verdict on whether each sample is real or fake.}
    \label{fig:model_outputs}
\end{figure*}

In addition to quantitative results, we conducted qualitative analyses to assess the interpretability of our model's decisions. Figure \ref{fig:fake-or-real} and \ref{fig:model_outputs} illustrates verdicts generated by TruthLens, highlighting interpretibility that contributed to the classification. These visualizations demonstrate the framework's ability to identify subtle artifacts in fake images.

\subsection{Discussion}
The success of ChatUniVi and Large Language Models (LLMs) within the TruthLens framework lies in their ability to integrate and reason across visual and textual modalities effectively. ChatUniVi excels by unifying image and video tokens into a shared representation, enabling a holistic understanding of visual artifacts and their contextual significance. This unified processing enhances the detection of subtle patterns, such as lighting inconsistencies and unnatural textures, which are often overlooked by traditional models. When paired with the reasoning capabilities of LLMs, ChatUniVi elevates detection accuracy and provides interpretable justifications for its decisions. The natural language explanations foster transparency and user trust, addressing critical challenges in high-stakes domains like media integrity and ethical AI. Furthermore, the adaptability of this framework across diverse datasets, such as ProGAN and Latent Diffusion Models (LDM), demonstrates its robustness and generalization capabilities, as evidenced by its superior AUC and classification accuracy metrics.

The training-free paradigm of TruthLens offers significant scientific and practical advantages. By eliminating the dependency on task-specific training, this approach ensures adaptability to emerging generative techniques without requiring large annotated datasets. This paradigm accelerates deployment and reduces computational overhead, making it highly scalable for various applications. Moreover, the reliance on pre-trained state-of-the-art models, like ChatUniVi and LLaVA, leverages their extensive training on diverse data, enabling the framework to detect artifacts in novel scenarios with minimal resource usage. Crucially, this paradigm shifts the focus to interpretable outputs by reframing detection as a Visual Question Answering (VQA) task, offering detailed natural language justifications that enhance trust and accountability. These attributes position TruthLens as a transformative solution for combating synthetic media, combining cutting-edge detection performance with transparency and user-centric design.



\section{Conclusion}\label{sec:conclusion}

TruthLens demonstrates promising results in fake image detection with interpretability. Our framework bridges the gap between detection accuracy and transparency, providing a robust tool for addressing challenges posed by synthetic content. By leveraging state-of-the-art vision-language and language models for testing, TruthLens showcases its flexibility and effectiveness across diverse datasets and scenarios. Its transparent decision-making process ensures human-readable explanations, promoting trust in its outputs. 
Future work includes extending the framework to handle emerging synthetic media types and exploring more advanced interpretability techniques to enhance user understanding. TruthLens represents a step forward in creating tools that are not only accurate but also explainable and adaptable to real-world applications.



{
    \small
    \bibliographystyle{IEEEtran}
    \bibliography{main}
}

\clearpage
\onecolumn

\end{document}